\documentclass[preprint,12pt]{elsarticle}

\RequirePackage{amsthm,amsmath,amsfonts,amssymb}
\RequirePackage{natbib}
\RequirePackage[colorlinks,citecolor=blue,urlcolor=blue]{hyperref}
\RequirePackage{graphicx}
\usepackage{caption} 
\usepackage{rotating}
\usepackage{subfig}
\usepackage{multirow}
\usepackage{xcolor}
\usepackage{bbm}
\usepackage{algorithm}
\usepackage{algpseudocode}
\usepackage[margin=1in]{geometry}

\newtheorem{theorem}{Theorem}[section]

\theoremstyle{remark}
\newtheorem{definition}[theorem]{Definition}

\journal{Transportation Research: Part B}

\begin{document}

\begin{frontmatter}



\title{Automatic Identification of Driving Maneuver Patterns using a Robust Hidden Semi-Markov Models}


\author[A]{Matthew Aguirre},
\author[B]{Wenbo Sun}
\author[A]{Jionghua (Judy) Jin}
\author[C]{Yang Chen}

\affiliation[A]{organization = {Department of Industrial and Operations Engineering, University of Michigan}, addressline = {1891 IOE Building 1205, Beal Ave}, city = {Ann Arbor}, postcode = {48109}, state = {MI}, county = {USA}}
\affiliation[B]{organization = {University of Michigan Transportation Research Institute}, addressline = {2901 Baxter Rd}, city = {Ann Arbor}, postcode = {48109}, state = {MI}, county = {USA}}
\affiliation[C]{organization = {Department of Statistics, University of Michigan}, addressline = {1085 S University Ave}, city = {Ann Arbor}, postcode = {48109}, state = {MI}, county = {USA}}





\begin{abstract}
There is an increase in interest to model driving maneuver patterns via the automatic unsupervised clustering of naturalistic sequential kinematic driving data. The patterns learned are often used in transportation research areas such as eco-driving, road safety, and intelligent vehicles. One such model capable of modeling these patterns is the Hierarchical Dirichlet Process Hidden Semi-Markov Model (HDP-HSMM), as it is often used to estimate data segmentation, state duration, and transition probabilities. While this model is a powerful tool for automatically clustering observed sequential data, the existing HDP-HSMM estimation suffers from an inherent tendency to overestimate the number of states. This can result in poor estimation, which can potentially impact impact transportation research through incorrect inference of driving patterns. In this paper, a new robust HDP-HSMM (rHDP-HSMM) method is proposed to reduce the number of redundant states and improve the consistency of the model's estimation. Both a simulation study and a case study using naturalistic driving data are presented to demonstrate the effectiveness of the proposed rHDP-HSMM in identifying and inference of driving maneuver patterns.
\end{abstract}


\begin{highlights}
\item A robust HDP-HSMM is proposed which produces more consistent results than the HDP-HSMM
\item An algorithm is described as to combat the inconsistency issues that arise from using an HDP prior
\item A simulation study is performed to show the impact of the proposed robust HDP-HSMM versus the basic HDP-HSMM in terms of parameter convergence and data segmentation
\item Real kinematic data is used to further compare robust HDP-HSMM and the basic HDP-HSMM in terms of learned maneuver patterns 
\end{highlights}


\end{frontmatter}


\section{Introduction}
\label{sec:intro}
\begin{figure}[!htb]
\centering
    \subfloat[]{\label{fig:Ex_Story_0_Map}{\includegraphics[trim={2cm 0 2cm 0},clip,width=0.4\textwidth]{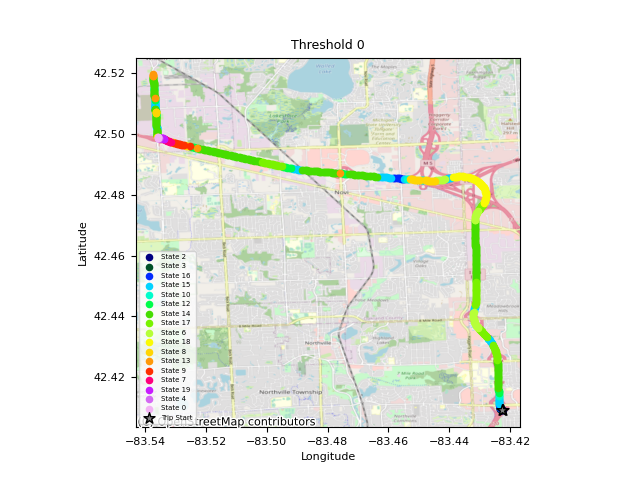}}}\hfill 
    \subfloat[]{\label{fig:Ex_Story_0_Signals}{\includegraphics[trim={0.5cm 0 0 0},clip,width=0.6\textwidth]{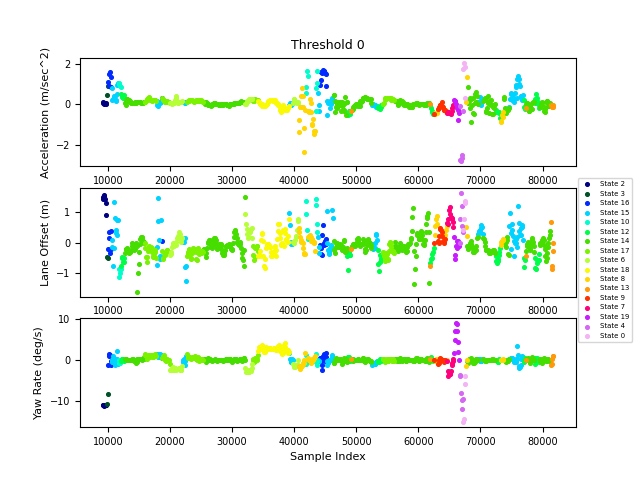}}}\hfill
\caption{An example trip and the kinematic signals belonging to it. Learned states from an HDP-HSMM are color coded as labels.}
\label{fig:Ex_Story_0}
\end{figure}

The analysis of vehicle driving styles is prominent to the field of intelligent transportation and vehicle calibration \citep{zhao2017trafficnet, rahman2017evaluation}. The term \textit{driving style} can be referred as a set of dynamic activities or steps that
a driver uses when driving. Hence, this type of research impacts eco-driving, road safety, and intelligent vehicles \citep{di2013stochastic, sagberg2015review, martinez2017driving}. To model these driving styles, one popular approach is the use of a Hierarchical Dirichlet Process Hidden Semi-Markov Model (HDP-HSMM) \cite{wang2018driving}. This model is powerful in that it considers the sequential nature of driving kinematic signals, and estimates data segmentation, behavior state duration, and state transition probabilities. The HDP-HSMM provides
semantical way for analyzing driver behaviors, and is thus popularly used for describing driving styles. Figure~\ref{fig:Ex_Story_0_Signals} shows an exemplar set of sequential kinematic signals belonging to the trip observed in Figure \ref{fig:Ex_Story_0_Map}. The signals are color-coded to reference a state segmentation determined by a HDP-HSMM. 

While the HDP-HSMM is powerful, literature outside of the field of transportation details how the model's use of an HDP prior can lead to redundant and inconsistent state estimations. This detail is important as it needs to be considered by researchers attempting to utilize the HDP-HSMM to describe driving styles. For example, Figure \ref{fig:Ex_Story_0} clearly has redundant states as seen by the green shaded states. The redundant states can make analysis of HDP-HSMM outputs across multiple datasets difficult for researchers hoping to utilize the HDP-HSMM to model driving styles. This paper addresses this issue by presenting an algorithm that reduces redundant states to improve consistency while still aligning to the structure of a basic HDP-HSMM. The presented algorithm results a more robust HDP-HSMM (rHDP-HSMM) that is expected to output a more consistent data segmentation, behavior state duration, and state transition probabilities than a basic HDP-HSMM. This will impact the transportation field in that driving maneuver patterns can be better grouped together for classification or behavioral studies.

The remainder of this paper is as follows. Section \ref{sec:background} will provide the background about HDP-HSMM's from a statistical perspective, and highlight the current set of approaches towards addressing the issues derived from the HDP prior. Section~\ref{sec: Basis of HDP-HSMMs} will provide the data description and the model formulation of a basic HDP-HSMM. Section~\ref{sec: rHDP-HSMM Methodology} discusses the details of inference for a HDP-HSMM, and how this paper's algorithm can be included within the inference to produce a more robust HDP-HSMM. Section~\ref{Case Study: Simulation} presents a simulation study, in which the rHDP-HSMM is compared to the basic HDP-HSMM based on simulated data. Section~\ref{Case Study: Naturalistic} presents a case study that uses realistic, naturalistic driving data to compare the rHDP-HSMM with the original HDP-HSMM method on the basis of describing driving patterns. Finally, Section \ref{Conclusion} summarizes new contributions and major conclusions of the paper.

\section{Background}
\label{sec:background}
\begin{figure}[!htb]
\centering
    \subfloat[HMM]{\label{fig: HMM Picture}{\includegraphics[width=0.5\textwidth]{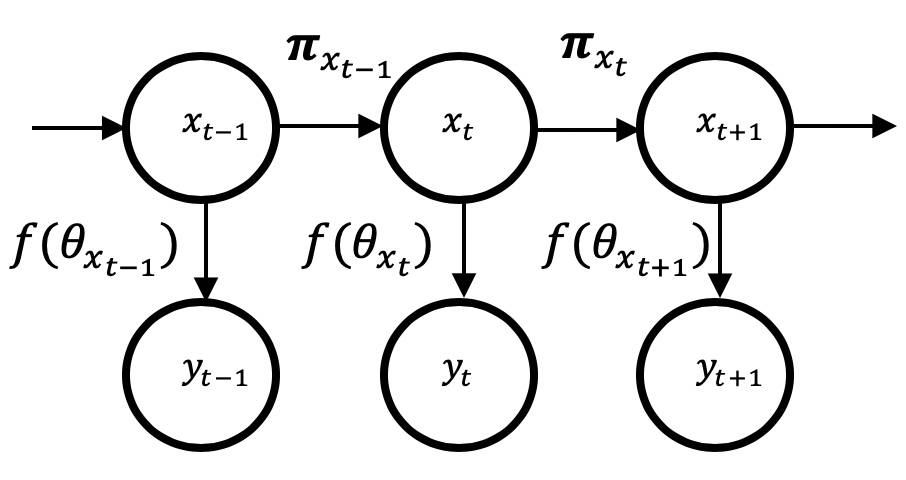}}}
    \subfloat[HSMM]{\label{fig: HSMM Picture}{\includegraphics[width=0.5\textwidth]{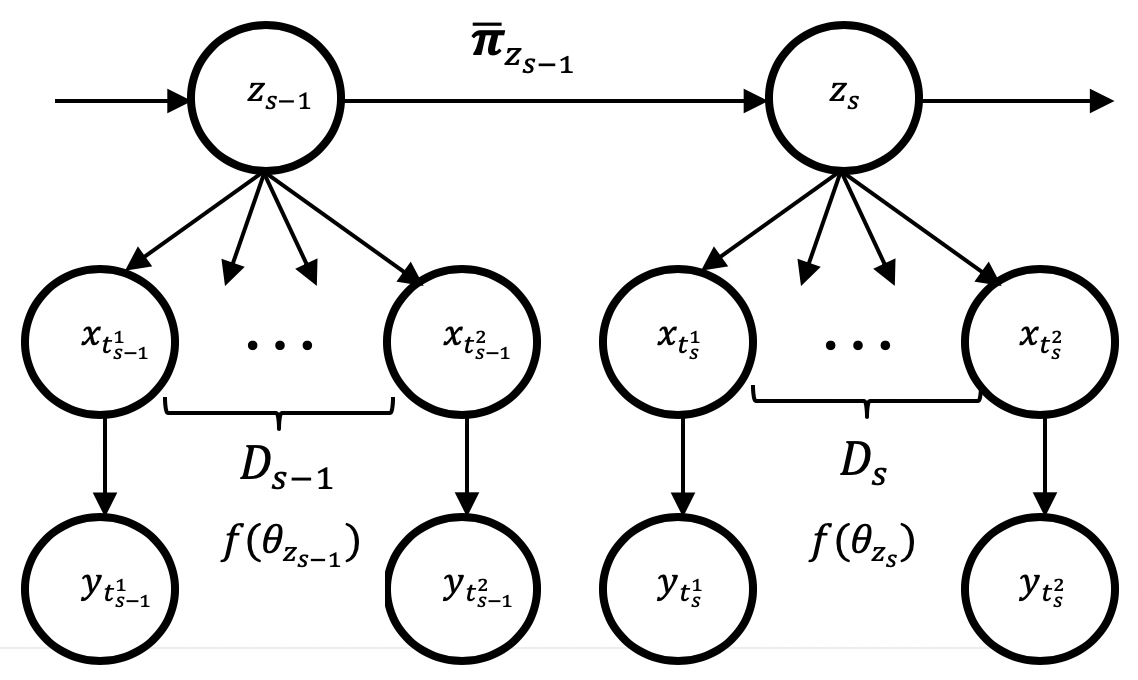}}}
\caption{A comparison between the structure of a Hidden Markov Model (HMM) and a Hidden Semi-Markov Model (HSMM). The variables and their descriptions are as follows: $x_t$ (hidden state at time $t$), $y_t$ (observed data at time $t$), $\pi_x$ (transition probabilities of state $x$), $f(\theta_x)$ (probability distribution of state $x$), $z_s$ (state of segment $s$), $D_s$ (state duration of segment $s$).}
\label{fig: HMM vs HSMM Structure}
\end{figure}

The HDP-HSMM was designed to improve upon the structure of a discrete state-space Hidden Markov Model (HMM). HMM's are also popularly used for describing sequential data \citep{fox2011sticky, wooters2007icsi,kolter2011redd, kim2011unsupervised,symul2021labeling,kamson2019multi}.
In particular, the HMM \citep{shumway2017time,rabiner1986introduction} utilizes a two-layer structure (Figure~\ref{fig: HMM Picture}) to represent sequential data observed at equally spaced time points. In this model, data is assumed to be generated from a set of probability distribution functions dependent on corresponding hidden states. The hidden states determine the data segmentation. Transitions among hidden states are modeled as a Markov Chain. This allows for the consideration of time sequence information during inference and further aids in the prediction of future states. One condition of using the Markov Chain is that the state duration of each hidden state is assumed to be Geometrically distributed. 

While the HMM is able to define data segmentation and state transitions, its definition of state duration is severely limited by the model's structure. This limitation lead to the development of the Hierarchical Dirichlet Process Hidden Semi-Markov Model (HDP-HSMM) \citep{johnson2013bayesian} which provided two key improvements to the HMM. The first improvement was the removal of the HMM's assumption of geometrically distributed state duration. As the HDP-HSMM uses a Semi-Markovian approach to model the state transitions $\bar\pi_{z_s}$, this removes self-transitions from the transition matrix. As a consequence, this frees the geometric distribution restriction on the duration $D_s$, which leads to a three-layer structure model as shown in   (Figure~\ref{fig: HSMM Picture}). In other words, users can choose different models for representing state duration, while allowing the segmentation of hidden states to be directly represented by $z_s$. 

The second improvement was the introduction of Dirichlet Processes to the model. The Dirichlet processes is an extension to the Dirichlet distribution, as atoms can be sampled from it based on an input distribution. However, one key difference is that the Dirichlet Process assigns a probability of drawing a new atom from the input distribution and a separate probability of drawing an atom based on the atoms seen in previous samples. The resulting distribution is discrete and similar to the input distribution, but also has the possibility of having infinite discrete atoms if infinite samples were drawn. This phenomenon is interesting in the context of HMMs and HSMMs, as the Dirichlet process can be used as a prior to the state transition probability vector \citep{beal2002infinite,Teh_2006_website_citation,johnson2013bayesian}. Doing this allows the probability vector length (i.e. models' number of states) to grow without limit during inference, which implies the Dirichlet process also acts like a prior on the number of clusters. In the HDP-HSMM, a Hierarchical Dirichlet Process (HDP) is used as a prior on the state transitions, which allows all the state transition probabilities to share a similar base distribution. This is beneficial, as all the states represented in the base distribution are shared between all the different state transition probabilities, while allowing each transition probability be dependent on the exit state. Hence, for the context of modeling of driving maneuvers, the HDP-HSMM is preferred as it allows greater flexibility in defining the relationship between the data and segmentation, state duration, and state transitions. 

While the Dirichlet Process's clustering properties have been seen as a tool to address the model selection for Bayesian nonparametric approaches \citep{teh2010dirichlet, jordan2010bayesian}, the Dirichlet Process is known to have inconsistency issues regarding estimation of the true number of states. \cite{miller2013simple} provided an example for Dirichlet Process Mixture Models which demonstrates how the posterior does not concentrate at the true number of components, and instead introduces extra clusters even if they are not needed. Under the context of HMMs, \cite{jordan2014gentle} showed how the Dirichlet Process also leads to the creation of redundant states, which presents an unrealistic rapid switching between states in the inferred transition matrices. Under the context of HSMM's, Figure \ref{fig:Ex_Story_0} shows how this side effect occurs even in the HDP-HSMM. However, for the HDP-HSMM, the redundancy issue also affects the inference of transition probabilities and duration estimation. 

A few works exist that focus on solving this issue for HMM's. \cite{gassiat2014posterior} discussed HMM's utilizing a Dirichlet prior, and the assumptions on the prior required for the consistency. \cite{fox2008hdp} developed the \textit{sticky} HDP-HMM (sHDP-HMM) to consider the issue of redundant states. This model adds a bias to the prior on the rows of the transition matrix which emphasizes self-transitions. This results in an increased state duration for each learnt state, which allows the sHDP-HMM to avoid redundant states with short state duration. However, this strategy cannot be applied to HDP-HSMM as the modeling structure of HMM's is inherently different from HSMM's. 
Outside of HMM and HSMM modeling, 
\cite{guha2019posterior} focused on the Dirichlet Process Mixture model, and presented the Merge-Truncate-Merge algorithm, which guaranteed a consistent estimate to the number of mixture components. This post-processing procedure takes advantage of the fact that the posterior sample tends to produce a large number of atoms with small weights, and probabilistically merges atoms together. 

Given these approaches, this paper attempts to address the HDP's inconsistency problem by taking inspiration from both the sticky HDP-HMM and the Merge-Truncate-Merge algorithm. The idea is to apply a merging procedure during inference which promotes longer durations and the avoidance of redundant states. In doing so, this paper's contribution will include demonstrating how the HDP-HSMM becomes robust to the inconsistencies brought by the HDP prior and how this paper's method can reduce the number of redundant states to better define driving maneuvers existing in Figure \ref{fig:Ex_Story_0_Map}. A brief summary, which describes where our model fits in relation to the other models described in HMM literature, is given in Table~\ref{tbl: rHDP_HSMM comparison}.

\begin{table}[ht]
\centering
\small
\begin{tabular}{|r|c|c|}
\hline
\textbf{\begin{tabular}[c]{@{}r@{}}State Duration \\ Distribution\end{tabular}} & \textbf{Model}                                            & \textbf{\begin{tabular}[c]{@{}c@{}}Extension \\ (not sensitive to prior)\end{tabular}} \\ \hline
\textbf{Geometric}                                                             & \begin{tabular}[c]{@{}c@{}}HDP-HMM\\ \citep{rabiner1986introduction}\end{tabular}  & \begin{tabular}[c]{@{}c@{}}sticky HDP-HMM\\ \citep{fox2011sticky}\end{tabular}                        \\ \hline
\begin{tabular}[c]{@{}c@{}}\textbf{Any Discrete}\\ \textbf{Distribution}\end{tabular}                                                                   & \begin{tabular}[c]{@{}c@{}}HDP-HSMM\\ \citep{johnson2013bayesian} \end{tabular} & \begin{tabular}[c]{@{}c@{}}robust HDP-HSMM\\ (This paper) \end{tabular}  \\ \hline            
\end{tabular}
\caption{\textit{Comparison of various HMM-based models versus our proposed robust HDP-HSMM (rHDP-HSMM).}}
\label{tbl: rHDP_HSMM comparison}
\end{table}

\section{Problem Formulation}
\label{sec: Basis of HDP-HSMMs}
\subsection{Data Description}
\label{sec: Data Description}
In this paper, a sequential dataset consists  of a series of observations collected at $T$ chronologically ordered time points. At each time point $t$, $y_t\in \mathbb{R}^p$ represents the $p-$dimensional signal responses. The sequential data is assumed to follow multiple phases; there exists a partition $1=t_1^1\leq t_2^1 \leq ... \leq t_S^1=T-D_S$, such that the elements within the $s-$th segment, denoted by $y_{t_{s}^{1}:t_{s}^{2}}$, are independent and identically distributed (i.i.d.) for a state duration of $D_s\in 1,2,\dots,S$. 

The objective of the data analysis is generalized to (1) identify distributional patterns that describe each phase, (2) identify the time duration distribution corresponding to each segment, and (3) identify the probability of transitioning from one distribution to another. The challenge lies in little information being available relating to the number of states, the states' durations, and the transition probability matrix.

\subsection{Basis of HDP-HSMMs and Notations}
\label{sec: HDP-HSMM Notation}

The HDP-HSMM accomplishes this objective with the following structure. The multivariate sequential data is represented by the sequence $(y_t)_{t=1:T}:=\{ y_t\in\mathbb{R}^p:  t=1,...,T \}$ and is assumed to transit among $K$ different hidden states. The hidden states at each time point $t$ are represented by the sequence $(x_t )_{t=1:T}:=\{ x_t\in\{1,2,\dots,K\}:  t=1,...,T \}$, and can be further divided into $S$ segments. Within each data segment $s\in \{1,2,\dots,S\}$, all hidden states share the same index (labeled by the super-state $z_s\in \{1,2,\dots,K\}$), and the state duration of the segment is denoted by $D_s$. As such, the start and end times of each segment $s$ are indexed by time stamps $t_s^1$ and $t_s^2$, respectively. They can be calculated as $t_s^1=\sum_{\bar{s}<s}D_{\bar{s}}$ and $t_s^2=t_s^1+D_s-1$ where $\bar{s}$ represents all the segments before segment $s$. The state of segment $s$ is assumed to be Markovian with a transition probability $\pi_{i,j}=\Pr(z_s=j\mid z_{s-1}=i)$, where the rows of the transition matrix are denoted as $\pi_i=[\pi_{i,1}\ \pi_{i,2}\ \dots\pi_{i,K}]$. However, as each state has a random state duration $D_s\sim g(\omega_{z_s})$, the HSMM does not permit self-transitions to occur. To consider this, the transition rows of $\pi_i$ are adjusted to $\bar{\pi}_i$ with each element being $\bar{\pi}_{i,j}=\frac{\pi_{i,j}}{1-\pi_{i,i}}(1-\delta_{i,j})$ (where $\delta_{i,j}=1$ if $i=j$; $\delta_{ij}=0$ otherwise).

The relationship between the observation sequence and the segmentation described above can be seen by the emission distribution functions $f(\theta_{z_s} )$ and the state duration probability mass functions $g(\omega_{z_s})$ with parameters $\theta_{z_s}$ and $\omega_{z_s}$ being dependent on segment $s$. The priors on $\theta_{z_s}$ and $\omega_{z_s}$ are denoted by $H$ and $G$ respectively. 

A Hierarchical Dirichlet Process (HDP) is used to define a prior on the rows of the transition matrix ($\pi_i$) to learn the number of unknown states. The HDP creates a countably infinite state-space and utilizes a stick-breaking process $\beta\sim\text{Beta}(\gamma)$ \citep{sethuraman1994constructive} to determine the number of unknown states ($K$). A smaller $\gamma$ ($\gamma$$\geq 0$) yields more concentrated distributions, which plays a part in shaping the transition pattern. Each row of the Markovian transition probability matrix is sampled from a Dirichlet process ($\pi_i\stackrel{\text{iid}}{\sim} \text{DP}(\alpha,\beta)$) and its similarity to the stick-breaking process depends on the concentration parameter $\alpha\in(0,\infty)$. 

The HDP-HSMM is shown in Figure \ref{fig: HSMM Picture} and can be formulated as follows:
\begin{equation}
   \begin{matrix}%
    \beta\sim \text{Beta}(\gamma), & & \\
    \pi_i\stackrel{\text{iid}}{\sim}\text{DP}(\alpha,\beta) & (\theta_i,\omega_i)\stackrel{\text{iid}}{\sim}H\times G & i=1,2,\dots,\\
    z_s\sim\bar{\pi}_{z_{s-1}}& & \\
    D_s\sim g(\omega_{z_s}) & &  s=1,2,\dots,\\
    x_{t_s^1:t_s^2}=z_s, & & \\
    y_{t_s^1:t_s^2}\stackrel{\text{iid}}{\sim}f(\theta_{z_s}) &  t_s^1=\sum_{\bar{s}<s}D_{\bar{s}} & t_s^2=t_s^1+D_s-1.
    \end{matrix} 
\end{equation}
Typically, Gibbs sampling approaches are used for statistical inference of the model parameters of the HDP-HSMM, which requires the full conditional distributions of the model parameters \citep{gelman2013bayesian}. 
The details of the general Gibbs sampling procedure and how this paper applies a merging algorithm within it to create a robust HDP-HSMM is presented in the next section.

\section{Proposed Robust HDP-HSMM}
\label{sec: rHDP-HSMM Methodology}
\subsection{Inference}
The details of the block sampling procedure presented in \cite{johnson2013bayesian} to infer the parameters for the HDP-HSMM are discussed here. Additional insight regarding this paper's proposed changes will also be included in this section. Assume initial values have been set for the state sequence, the emission parameters, the duration parameters, and the transition probabilities:
$$(x_t )^{(0)}, \{\theta_i\}^{(0)},\{\omega_i \}^{(0)},\{\pi_i \}^{(0)}.$$

\textbf{Step 1:} The block sampling procedure begins iteration $m =1$ with the sampling of the emission, duration, and transition distribution parameters. The distributional parameters can be sampled independently of one another, conditional on data assigned to each state $i$ under the current state sequence $(x_t )^{(m-1)}$. Assuming distributions with conjugate priors are utilized within the HDP-HSMM, this step can be simplified significantly into the following statement:
\begin{align*}
\{\theta_i\}^{(m)}	&\sim	h_{\theta_i} (\theta_i|(x_t )^{(m-1)},(y_t),H,G,\beta)\\
\{\omega_i\}^{(m)}	&\sim	h_{\omega_i} (\omega_i|(x_t )^{(m-1)},(y_t),H,G,\beta)\\
\{\pi_i\}^{(m)}	&\sim	h_{\pi_i} (\pi_i|(x_t )^{(m-1) },(y_t ),H,G, \beta),
\end{align*}
where $h_\theta$ refers to the updated posterior corresponding to the conditional distribution with parameter $\theta$. 

\textbf{Step 2:} Once a new set of parameters have been sampled, it is practical to apply some identifiability constraints to the parameters to help ensure state switching does not occur during the sampling procedure. State switching is a problem mentioned in literature  \citep{jasra2005markov, sperrin2010probabilistic}, in which the permutation of defined states is not considered during the sampling procedure. Identifiability constraints ensure the order of states does not change between iterations of the sampling procedure, and helps ensure the posterior chain is not multimodal at the end of the sampling procedure. While many types of constraints can be applied, such as rearranging the states such that $\theta_1 < \theta_2 < \theta_3<\dots$, the constraints used in this paper are be mentioned in each section directly.

\textbf{Step 3:} After identifiability constraints have been applied, the new state sequence can be sampled. \cite{johnson2013bayesian}'s procedure makes use of the following backwards messages:
\begin{align*}
    B_t(i):=&p(y_{t+1:T}|x_t=i,F_t=1)\\
    =&\sum_j B_t^*(j)p(x_{t+1}=j|x_t=i)\\ 
    B_t^*(i):=&p(y_{t+1:T}|x_{t+1}=i,F_t=1)\\
    =&\sum_{d=1}^{T-t} B_{t+d}(i)p(D_{t+1}=d|x_t=i)p(y_{t+1:t+d}|x_{t+1}=i,D_{t+1}=d)\\
    &+p(D_{t+1}>T-t|x_{t+1}=i)p(y_{t+1:T}|x_{t+1}=i,D_{t+1}>T-t)\\
    B_{T}(i):=&1,
\end{align*}
where $F_t=1$ denotes a new segment begins at $t+1$, and $D_{t+1}$ denotes the duration of the segment that begins at time $t+1$ \citep{murphy2002hidden}.
The procedure for obtaining the posterior state sequence begins by drawing a sample for the first state using the following formula:
\begin{align*}
    p(x_1=k|y_{1:T})\propto p(x_1 = k)B_0^*(k).
\end{align*}
Next, a sample is drawn from the posterior duration distribution by conditioning on sampled initial state $\bar x_1$:
\begin{align*}
    p(D_1 = d|y_{1:T},x_1=\bar x_1, F_0=1) &= \frac{p(D_1 = d)p(y_1:d|D_1=d,x_1 = \bar x_1, F_0 = 1)B_d(\bar x_1)}{B_0^*(\bar x_1)}.
\end{align*}
The rest of the state sequence can be sampled assuming the new initial state has distribution $p(x_{D_1+1}=i|x_1= \bar x_1)$ and repeating the process, until a state is assigned for all indices $t=1,\dots,T$.

\textbf{Step 4:} Once the new state sequence is sampled, the Gibbs sampling procedure normally returns back to Step 1, increments $m$ by 1, and repeats Steps 1 to 3 until posterior convergence. However, before doing that, this paper propose adding an additional sampling Step 4 that removes redundant states from the posterior state sequence

\begin{equation}
\label{eq: step 4}
(\tilde x_t)^{(m)}	\sim	h_{(\tilde x_t)} ((\tilde x_t)|\{\theta_i\}^{(m)},\{\omega_i\}^{(m)},\{\pi_i\}^{(m)}(x_t )^{(m)},(y_t),H,G,\alpha),
\end{equation}
where $h_{(\tilde x_t)} (\cdot)$ represents a sampling step proposed by this paper to promote robustness.

\subsection{Implementation of Step 4}
The proposed Step 4 is the main contribution of this paper. This section will provide the details on how to implement Equation \ref{eq: step 4} described in Step 4 above. The procedure is described by first defining redundancy between two states:

\begin{definition}
\label{Def: Divergence}
\textit{In the state sequence $(x_t)_{t=1:T}$, the states $i$ and $j$ are identified as \textbf{redundant states} if $\mathcal{D}\big(f(\theta_i),f(\theta_j)\big)\leq \tau$, where $\tau$ is the decision threshold and $\mathcal{D}\big(f(\theta_i),f(\theta_j)\big)$
is a measure of divergence that gets larger when the distributions $f(\theta_i)$ and $f(\theta_j)$ are more different from one another.}
\end{definition} 
Although $\mathcal{D}\big(f(\theta_i),f(\theta_j)\big)$ can be any measure of divergence satisfying Definition \ref{Def: Divergence}, the remainder of the paper will assume $\mathcal{D}\big(f(\theta_i),f(\theta_j)\big)= ||(\theta_i - \theta_j)||_2$ is the $\ell^2$ norm of the difference in parameters. 

Now that redundancy has been defined, the details of Equation \ref{eq: step 4} can be represented by Algorithm \ref{alg:stateseq alg}. In short, the procedure samples a new state sequence that contains no redundant states. \cite{johnson2013bayesian} describes a weak-limit approximation to the Dirichlet Process prior,
\begin{align*}
\beta | \gamma &\sim \text{Dir}(\gamma,\dots,\gamma)\\
\pi_j | \beta &\sim \text{Dir}(\alpha\beta_1,\dots,\alpha\beta_K), \ \ j = 1,\dots,K,
\end{align*}
as well as an augmentation that introduces auxiliary variables which are added to the $\beta$ vector to preserve conjugacy. This approximation eases the use of sampling procedures when dealing Dirichlet Processes \citep{johnson2012dirichlet}. Taking this approach, the $\beta$ vector takes no consideration of redundant states, which may negatively impact the posterior of $\pi_j$. The presence of redundant states means the posterior transition probabilities contain extra transitions to and from redundant states, which dilute the underlying transition process. To counter this, $h_{(\tilde x_t)} (\cdot)$ aims to adjust the $\beta$ vector in this step as to discourage transitions to redundant states in future steps, and preserve the true underlying transition process.

Algorithm \ref{alg:stateseq alg} describes $h_{(\tilde x_t)} (\cdot)$ entirely. The procedure begins by initializing a new vector $\tilde\beta$, a new state sequence $(\tilde x_t)^{(m)}$, and taking the input of a similarity threshold $\tau$. Taking inspiration from \cite{guha2019posterior}, the states order is firstly randomized in which redundancy is checked. This is to ensure the start of the merging procedure begins at a point close to the ``central mass'' of the emission distribution clusters with a high probability. Going through the order, if the state exists within the new state sequence $(\tilde x_t)^{(m)}$, the algorithm proceeds to find similar states based on our similarity metric and similarity threshold. Weights are then defined which will determine the probability of retaining a state from the set of redundant states. These weights are determined by the probability of other non-similar states transitioning to the state of interest and then normalized. The state by which to retain is selected randomly in accordance to the probabilistic weights, and the rest of the similar states are erased from the state sequence. Vector $\tilde \beta$ is further updated by weakening the unselected similar states values in the vector.

\begin{algorithm}
\small
\caption{Sample a State Sequence Containing No Redundant States}\label{alg:stateseq alg}
\begin{algorithmic}
\State Initialize $\tilde\beta = \beta$, $(\tilde x_t)^{(m)} = (x_t)^{(m)}$, and define similarity threshold $\tau$
\State Reorder $\{\theta_i:i\in (\tilde x_t)^{(m)}\}$ into new order $\{\theta_{I_i}:i\in (\tilde x_t)^{(m)}\}$ using random sampling without replacement where 
\begin{itemize}
    \item $i$ corresponds to index the unique states existing in $(\tilde x_t)^{(m)}$
    \item $I_i$ corresponds to the new index of state $i$ in the new order $I = \{1,2,3,\dots\}$ 
\end{itemize}
\While{$I$ is not an empty set}
        \State Let $i$ correspond to the first $I_i$ appearing in the new order $I$
        
        \State Calculate $\mathcal{D}\big(f(\theta_i),f(\theta_j)\big)$ for all $j\neq i$ where $j\in (\tilde x_t)^{(m)}$ \Comment{Similarity metric.}
        \State Define set $J = \{j:\mathcal{D}\big(f(\theta_i),f(\theta_j)\big) 
        \leq \tau\}$ and set 
        $J' = \{j:\mathcal{D}\big(f(\theta_i),f(\theta_j)\big)>\tau\}$
        \For{$j\in J$}
            \State $\Pi_j= \sum_{i\in J'}\pi_{i,j}$ \Comment{Weights depend on transition probabilities from non-similar states.}
        \EndFor
        \State Sample $j^*$ from $\text{P}(j^*)$ where $\text{P}(j^*=j)=\Pi_j / (\sum_j \Pi_j)$ \Comment{$j^*$ is the redundant state to keep.}
        \State Update $\tilde \beta_j = 0.1 * \beta_j$ for all $j\in J$ where $j\neq j^*$ \Comment{Influence transition prior.}
        \State Update $\tilde x_t = j^*$ for all $\{t:\tilde x_t \in J\}$  \Comment{Influence data used for inference.}
        \State Remove $I_j$ from $I$ for all $j\in (J\cup i)$. \Comment{Prevent merging these states in future iterations.}
\EndWhile
\State Output final $\tilde\beta$ and $(\tilde x_t)^{(m)}$ \Comment{These will be used in next iteration of Gibbs sampling.}
\end{algorithmic}
\end{algorithm}

After implementing Algorithm \ref{alg:stateseq alg}, the sampling procedure is allowed to return to Step 1. Noticeably, every time this step is implemented, the algorithm begins with the originally sampled $\beta$ and $(x_t)^{(m)}$, but ends with a $\tilde \beta$ and $(\tilde x_t)^{(m)}$ that encourages the transition matrix in Step 1 to promote transitions to non-redundant states and allow larger sample sizes for the available emission posteriors. Mathematically, the only adjustments made to Step 1 that reflect this dependency is
\begin{align*}
\{\theta_i\}^{(m)}	&\sim	h_{\theta_i} (\theta_i|(\tilde x_t )^{(m-1)},(y_t),H,G,\tilde \beta)\\
\{\omega_i\}^{(m)}	&\sim	h_{\omega_i} (\omega_i|(\tilde x_t )^{(m-1)},(y_t),H,G,\tilde \beta)\\
\{\pi_i\}^{(m)}	&\sim	h_{\pi_i} (\pi_i|(\tilde x_t )^{(m-1) },(y_t ),H,G,\tilde \beta),
\end{align*}
which preserves the Markov Chain structure of a Gibbs Sampler.

\section{A Simulation Study}
\label{Case Study: Simulation}\
In this section, simulations are used to demonstrate the advantages of the proposed rHDP-HSMM method. The robustness and modeling accuracy is compared with the existing HDP-HSMM method. The simulation is designed as follows.

For each simulation, a sequence of observed data is generated with 30 total change points based on the distributions and parameters in Table \ref{tbl: True Params}. The emission parameters were specifically selected as they feature some small overlap between their distributions.
\begin{table}[ht]
\centering
\small
\begin{tabular}{|l|cc|c|ccc|}
\hline
                      & \multicolumn{2}{c|}{\textbf{Emission}}                 & \textbf{Duration} & \multicolumn{3}{c|}{\textbf{Transition}}                                                         \\ \hline
\textbf{Distribution} & \multicolumn{2}{c|}{\textbf{Normal}}                   & \textbf{Poisson}  & \multicolumn{3}{c|}{\textbf{N/A}}                                                                \\ \hline
\textbf{Parameter(s)} & \multicolumn{1}{c|}{\textbf{Mean}} & \textbf{Variance} & \textbf{Rate}     & \multicolumn{1}{c|}{\textbf{State 1}} & \multicolumn{1}{c|}{\textbf{State 2}} & \textbf{State 3} \\ \hline
\textbf{State 1}      & \multicolumn{1}{c|}{4}             & 1                 & 6                 & \multicolumn{1}{c|}{0}                & \multicolumn{1}{c|}{0.3}              & 0.7              \\ \hline
\textbf{State 2}      & \multicolumn{1}{c|}{0}             & 1                 & 6                 & \multicolumn{1}{c|}{0.8}              & \multicolumn{1}{c|}{0}                & 0.2              \\ \hline
\textbf{State 3}      & \multicolumn{1}{c|}{-4}            & 1                 & 6                 & \multicolumn{1}{c|}{0.4}              & \multicolumn{1}{c|}{0.6}              & 0                \\ \hline
\end{tabular}
\caption{\textit{The list of the true parameters in the hypothetical dataset with three different states.}}
\label{tbl: True Params}
\end{table}

\begin{figure*}[hbt!]
     \centering \includegraphics[scale = 0.45]{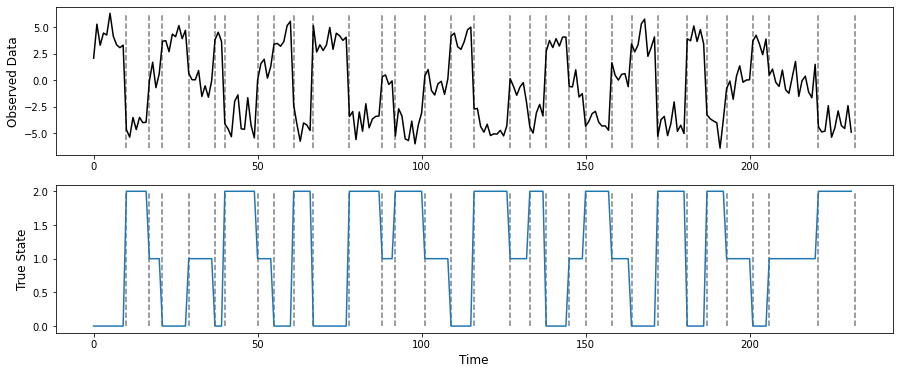} 
\caption{Example of simulated data based on Table \ref{tbl: True Params} and it's corresponding states.}
\label{fig: Sim_Data_ex}
\end{figure*}

The generated sequence begins with a state being randomly selected from the three listed in Table \ref{tbl: True Params}. A length of duration is sampled from the selected state's duration distribution, which determines how many samples to draw from that state's emission distribution. Once the emission samples are collected, they are stored in the sequence, and the next state is sampled according the to that state's transition probability. The process is repeated 30 times to create a simulated sequence of ``observed'' data. An example of a simulated dataset can be observed in the Figure \ref{fig: Sim_Data_ex}.

In each simulation, both the HDP-HSMM and the rHDP-HSMM are trained on the observed data with the same initial distributions and priors. The prior distributional forms were selected as to allow models to make use of conjugate relationships. Their parameters were selected as to ensure the true distributional parameters could be inferred with high probability. Each simulation's initial parameter values for the HDP-HSMM and rHDP-HSMM were drawn according to the selected prior. The maximum number of states for both models was set to 20. Each state's initial emission distribution was assumed $\text{Normal}(\mu, \sigma^2)$. The mean's prior distribution was set to $\mu\sim \text{Normal}(\mu_0 = 0, \sigma^2_0 = 4)$. The variance's prior distribution was set to $\sigma^2\sim \text{InvGamma}(a_0 = 2, b_0 = 2)$. The initial duration distributions were assumed $\text{Poisson}(\lambda)$, with prior $\lambda\sim\text{Gamma}(a_1=1,b_1=7)$. The transition distributions for each state is assumed to be $\pi_i\sim\text{Multinomial}(a_2)$, with the prior $a_2\sim\text{Dirichlet}(a_3=\mathbbm{1}^{20})$. Both models had identifiability constraints implemented such as to order their states in increasing order of the posterior mean of their emission distribution. Furthermore, both models performed their respective Gibbs procedure over a maximum of 10000 iterations, or until their Gelman-Rubin statistic \citep{gelman1992single} reached less than 1.1. The burn-in period for both models was set to 100 iterations. Every 5th iteration of the sampled parameter chains was collected as to remove autocorrelation (resulting in a chain of 2000 length if convergence was not met). The rHDP-HSMM threshold for removing redundant states was set to 1.5. The posterior parameter values for each state was calculated as the mean of the most recent 20\% of samples collected from the posterior parameter chains. The posterior sequence was selected to be the mode of the most recent 20\% of samples collected from the posterior state sequence.

\begin{figure}[htp!]
\centering
\subfloat[]{\label{fig:Sampled_Data_HDP_convergence_mean}{\includegraphics[width=\textwidth]{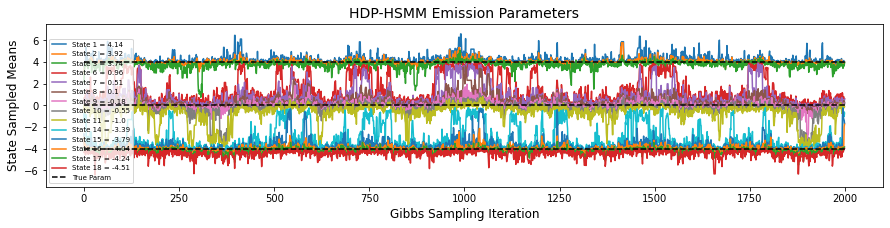}}}\hfill
\subfloat[]{\label{fig:Sampled_Data_rHDP_convergence_mean}{\includegraphics[width=\textwidth]{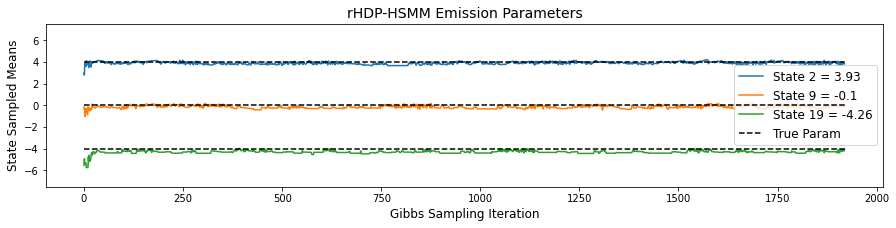}}}\hfill
\caption{HDP-HSMM versus rHDP-HSMM emission convergence on simulated data.}
\label{fig: sim: emission convergence}
\end{figure}

\begin{figure}[h]
\centering
\subfloat[]{\label{fig:Sampled_Data_HDP_convergence_dur}{\includegraphics[width=\textwidth]{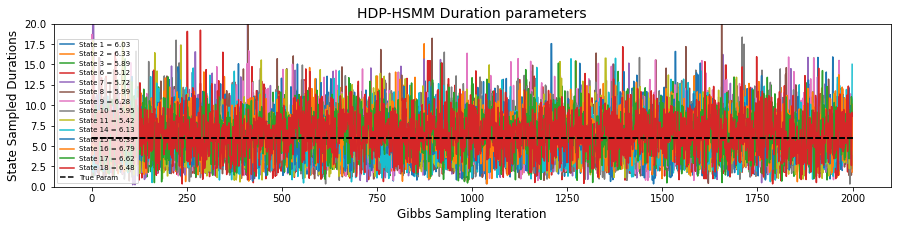}}}\hfill
\subfloat[]{\label{fig:Sampled_Data_rHDP_convergence_dur}{\includegraphics[width=\textwidth]{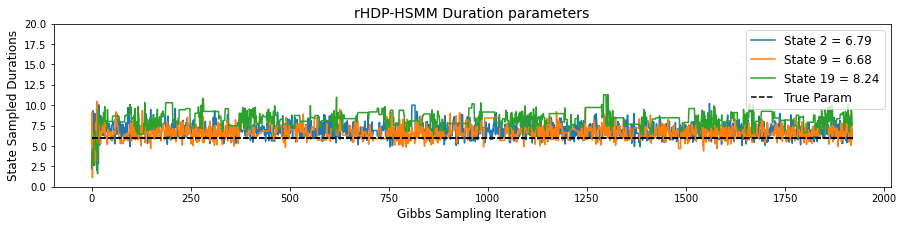}}}\hfill
\caption{HDP-HSMM versus rHDP-HSMM duration convergence on simulated data.}
\label{fig: sim: duration convergence}
\end{figure}

\begin{figure}[htp!]
\centering
\subfloat[]{\label{fig: sim: HDP-HSMM Stateseqs}{\includegraphics[width=\textwidth]{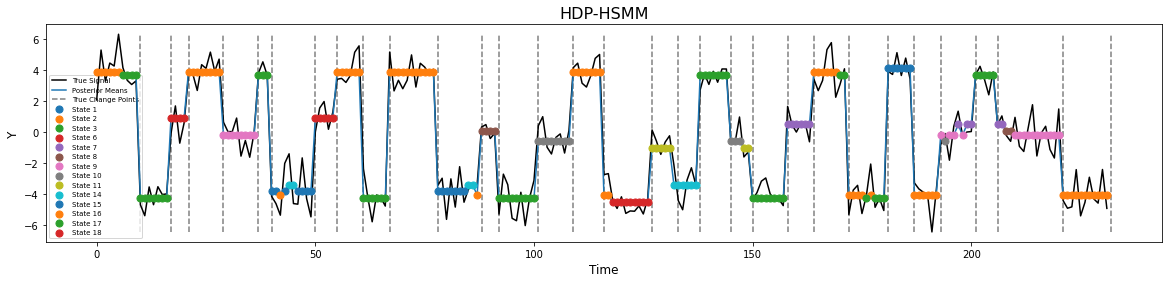}}}\hfill
\subfloat[]{\label{fig: sim: rHDP-HSMM Stateseqs}{\includegraphics[width=\textwidth]{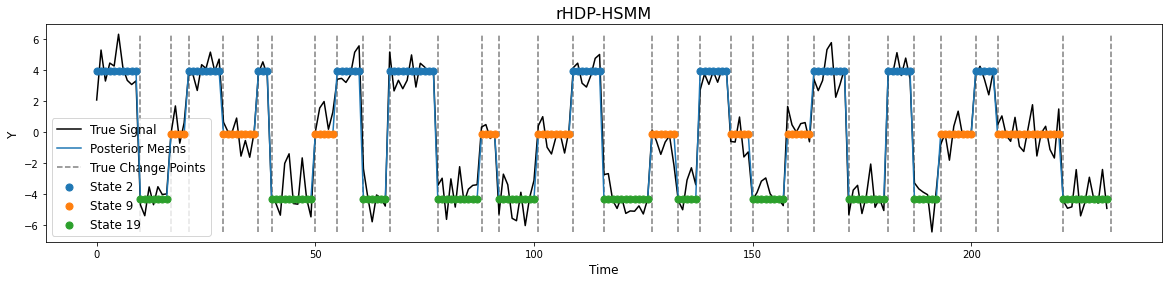}}}\hfill
\caption{HDP-HSMM versus rHDP-HSMM labeling of simulated data.}
\label{fig: sim: stateseqs comparison}
\end{figure}

The results of a single simulation are shown in Figures \ref{fig: sim: emission convergence}, \ref{fig: sim: duration convergence}, and \ref{fig: sim: stateseqs comparison}. Figure \ref{fig: sim: emission convergence} compares the HDP-HSMM and rHDP-HSMM's emission distribution convergence. The states shown in the plots are the states appearing in the final learned state sequence for each model. Each state is indicated by a different color. The true parameters are indicated by the dashed lines. While both models' posteriors are concentrated around the true parameters, the HDP-HSMM's posterior is multimodal for many states. Figure \ref{fig:Sampled_Data_HDP_convergence_mean} shows how the many states rapidly switch which true state they want to encapture across sampling iterations. With regards to the duration, Figure \ref{fig: sim: duration convergence} displays how both models posteriors are concentrated near the true duration. However, the variance of the HDP-HSMM's posterior samples is far larger than the variance of the rHDP-HSMM. This could be due to a large variation of samples being allocated to each state in the HDP-HSMM. To see this, Figure \ref{fig: sim: stateseqs comparison} shows the posterior state sequence estimated by both models. While the rHDP-HSMM distributes samples to each state under the constraint of removing redundant states, the HDP-HSMM's redundant states leave many states with very few samples left to estimate their duration. On a more positive note, both models are able to capture most of the true change points, however the HDP-HSMM leaves an impression of many more change point occurances. Meanwhile, rHDP-HSMM clearly separates each of the 3 states from one another and captures only the locations of the true change points in the data.

\begin{figure*}[hbt!]
     \centering \includegraphics[scale = 0.75]{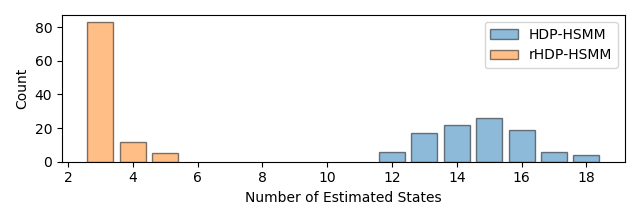} 
\caption{The number of estimated states from 100 simulations comparing both a HDP-HSMM and the rHDP-HSMM.}
\label{fig: Sim_num_States}
\end{figure*}

The simulation is repeated 100 times, and the results are shared in Figure \ref{fig: Sim_num_States} and Table \ref{tbl: Simulation Convergence Results}. Looking at the number of estimated states between the HDP-HSMM and the rHDP-HSMM, it is clear that the rHDP-HSMM's inference procedure removes states that would be otherwise present in a standard HDP-HSMM (Figure \ref{fig: Sim_num_States}). In fact, 80 of the 100 simulations resulted in the rHDP-HSMM correctly inferring the true number of states. Furthermore, Table \ref{tbl: Simulation Convergence Results} shows that the rHDP-HSMM converged on average with fewer iterations than the HDP-HSMM. This table also shows that while both models are able to correctly capture all the true change points, the standard HDP-HSMM tends to estimate many more change points than the rHDP-HSMM. This is due to the redundancy issue, which the rHDP-HSMM eliminates through its modified inference procedure.

\begin{table}[hbt!]
\centering
\small
\begin{tabular}{|r|c|c|}
\hline
                                        & \textbf{HDP-HSMM} & \textbf{rHDP-HSMM} \\ \hline
\textbf{Num. of Converged Simulations}     & 64                & 80                 \\ \hline
\textbf{Avg. Num. of Gibbs Iterations}     & 1372.5            & 993.5              \\ \hline
\textbf{Avg. Num. of Missed Change Points} & 1.0               & 1.3                \\ \hline
\textbf{Avg. Num. of Extra Change Points}  & 15.0              & 1.5                \\ \hline
\end{tabular}
\caption{\textit{The results of 100 simulations. Averaged values are calculated only from the iterations that converged.}}
\label{tbl: Simulation Convergence Results}
\end{table}

\section{A Case Study on Naturalistic Driving Data}
\label{Case Study: Naturalistic}
\begin{figure}[htp]
\centering
\subfloat[]{\label{fig:Ex_Story_half_Map}{\includegraphics[trim={0.5cm 0 0 0.5cm},clip,width=0.5\textwidth]{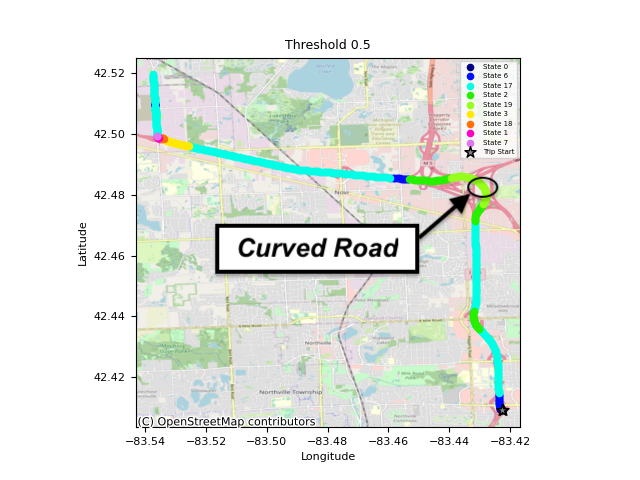}}}\hfill
\subfloat[]{\label{fig:Ex_Story_half_signals}{\includegraphics[trim={0.5cm 0 0 0},clip,width=0.5\textwidth]{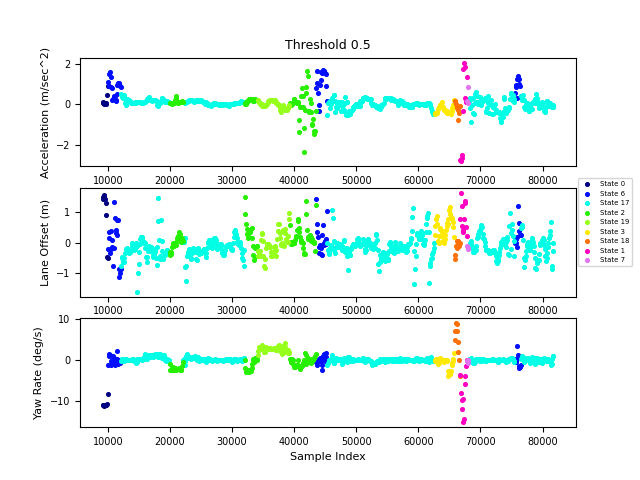}}}\hfill
\caption{A different segmentation of the road shown in Figure \ref{fig:Ex_Story_0} labeled by a rHDP-HSMM under a threshold of 0.5.}
\label{fig:Ex_Story}
\end{figure}

The benefit of the proposed rHDP-HSMM is demonstrated via the real-world application of modeling vehicle driving maneuver patterns. This type of modeling is useful for the development intelligent driving assistant systems and autonomous driving vehicles. The dataset analyzed in this study was collected by University of Michigan's Transportation Research Institute \citep{nodine2011integrated}. Several kinematic driving signals were collected from human-driven vehicles during their everyday activities. This naturalistic dataset is rich with information related to discover common driving maneuvers and behaviors. \citep{zhao2017trafficnet}. Signals are recorded on trip by trip basis, which begins when the vehicle is turned-on and ends when the vehicle is turned-off. An example of a trip can be seen in Figure \ref{fig:Ex_Story}. 

The kinematic signals of interest are acceleration, lane offset, and yaw rate. Acceleration and lane offset reflect a driver's intention of moving in the longitudinal and lateral directions respectively. Yaw rate captures a driver's intention of of changing the forward direction of the car. Together, they form a multivariate time-series sampled at 10 Hz which should be highly correlated with human-driving behaviors. An example of the collected signals is shown in Figure \ref{fig:Ex_Story}. As maneuvers are expected to switch at a low frequency, the original data is down-sampled to 1 Hz by averaging every 10 data points. 

Both the HDP-HSMM and a 0.5 threshold rHDP-HSMM are applied to trip shown in Figure \ref{fig:Ex_Story} under the following setup. A 3-dimensional multivariate Gaussian distribution is used for the emission distribution ($Y\sim \text{MVN}(\mu,\Sigma)$). The priors to the emission mean and variance are selected as
$$\mu \sim \text{MVN}([0,0,0], [[1,0,0],[0,1,0],[0,0,1]])$$
$$\Sigma \sim \text{Inverse-Wishart}(2, [[1,0,0],[0,1,0],[0,0,1]]).$$
Each state's duration is assumed Poisson distributed ($D\sim\text{Poisson}(\lambda)$) with the prior $\lambda\sim \text{Gamma}(a=1,b=7)$. The identifiability constraints are constructed as to arrange the states in the order of smallest to largest mean and duration. The maximum number of states was limited to $20$. The kinematic signals are normalized with respect to the signals observed during the trip. The learned emission means are transformed back to original space once the training is complete for analysis purposes. 

The colors in Figure \ref{fig:Ex_Story} represent the labeling results after training the 0.5 threshold rHDP-HSMM. Noticeably, the rHDP-HSMM segments the road into 9 states. Looking deeper at Figure \ref{fig:Ex_Story_half_signals}, it is clear that each state is primarily dictated by changes in yaw rate. Hence this model is able to capture portions of the road where various turning maneuvers are intended by the driver (Figure \ref{fig:Ex_Story_half_Map}). Comparing Figure \ref{fig:Ex_Story_half_Map} with the HDP-HSMM segmentation shown in Figure \ref{fig:Ex_Story_0_Map}, it is clear how the rHDP-HSMM merged the HDP-HSMM's 17 states into a more clear representation of maneuvers used on the road.

\begin{figure}[h]
\centering
\subfloat[]{\label{fig:Ex_Curved_0}{\includegraphics[width=\textwidth]{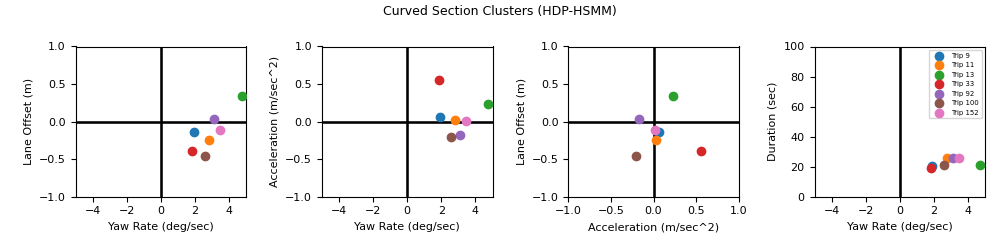}}}\hfill
\subfloat[]{\label{fig:Ex_Curved_half}{\includegraphics[width=\textwidth]{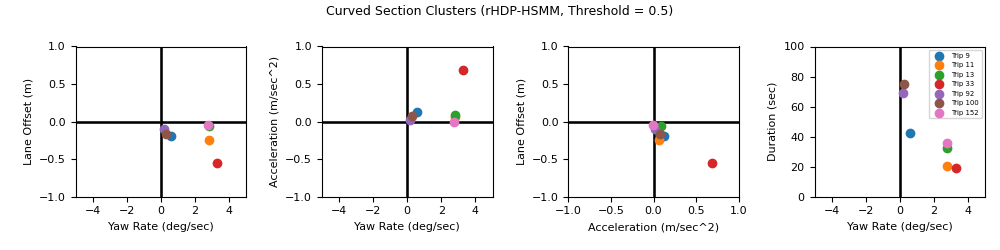}}}\hfill
\caption{Emission means corresponding to the kinematic signals from 7 different trips occuring on the curved portion of road shown in \ref{fig:Ex_Story}. Figure \ref{fig:Ex_Curved_0} shows the means from the original HDP-HSMM, while Figure \ref{fig:Ex_Curved_half} shows the means from the proposed rHDP-HSMM.}
\label{fig:Ex_Curved_Clusters}
\end{figure}

The rHDP-HSMM and HDP-HSMM are further compared in Figure \ref{fig:Ex_Curved_Clusters} by using states obtained from the curved portion of the road marked in Figure \ref{fig:Ex_Story_half_Map}. Six other trips existed where the same driver drove on that part of the road. Hence, both the HDP-HSMM and the rHDP-HSMM are trained again on each of the other trips under the same initial parameters. The learned states from each model which occurred on the marked portion are analyzed in Figure \ref{fig:Ex_Curved_Clusters}. Figures \ref{fig:Ex_Curved_0} and \ref{fig:Ex_Curved_half} shows the emission means and durations learned by the HDP-HSMM and the rHDP-HSMM respectively. Interestingly, Figure \ref{fig:Ex_Curved_half} shows how the rHDP-HSMM concentrates the emission means in various quadrants of the graph. These quadrants relay a positive yaw rate, a negative lane offset, and a positive acceleration in all the learnt means. The concentration of these means in each quadrant indicate a consistency in maneuvers among the various trips, which translates to a left turning action intended by the driver. This same conclusion is not easily recognizable in Figure \ref{fig:Ex_Curved_0}, as the HDP-HSMM loses this consistency in the learnt means. The difference in learning procedure between the HDP-HSMM and the rHDP-HSMM suggests that the HDP-HSMM's lack of concentrated means derives from the HDP-HSMM overestimating the number of states. As the rHDP-HSMM inference procedure merges similar states together, the emission means of each state can be inferred with a greater amount of data, providing both more consistent estimates and more consistent conclusions.

\section{Discussion and Conclusion}
\label{Conclusion}
The HDP-HSMM is a powerful model for discovering driving maneuver patterns from kinematic driving data. This paper details an extension to the HDP-HSMM in which this paper refers to as a robust HDP-HSMM (rHDP-HSMM). This model provides a solution to the inconsistency problem caused by the HDP prior. Looking through the lens of a weak-limit approximation of the HDP prior, the problem typically occurs as the Dirichlet distribution takes no consideration for redundant states, which dilutes the underlying transition process. The rHDP-HSMM solves this issue by adjusting the sample from Dirichlet distribution by checking which states can be merged together. The model then scales down the weights which encourage transitions to redundant states. As a result, the rHDP-HSMM learns fewer redundant states and estimates longer state durations when compared to the original HDP-HSMM. This change leads to improved segmentation and more accurate transition probability representation, which is useful for the application of learning driving maneuvers.

Two case studies are presented to further demonstrate the ability of the proposed rHDP-HSMM over the HDP-HSMM. The first study is a simulation which utilizes 1-dimensional normal distributions for the emission function. The rHDP-HSMM demonstrates a clear improvement with regards to the posterior chains. The emission parameters converge much faster, the duration posteriors have far less variance than the HDP-HSMM's duration posterior, and finally the posterior state sequence presents far less change points than the HDP-HSMM's. Over the course of 100 simulations, the rHDP-HSMM out performs the HDP-HSMM in terms of convergence and having less extra change points relative to the truth. 

The second study demonstrates of the effectiveness of the model in identifying and inferring driving maneuver patterns from a naturalistic dataset of kinematic signals. It is shown how the rHDP-HSMM's merging procedure reduces the number of states to describe a trip from 17 to 9 states when compared to a regular HDP-HSMM. The states are highly interpretable and now specifically capture portions of the road where various turning maneuvers are intended by the driver. In addition to this, the study also compares the results from multiple trips occurring on a curved portion of the road. The results show how the rHDP-HSMM consistently estimates similar emission distributions from multiple trips when compared to the original HDP-HSMM estimates.

In both studies, the rHDP-HSMM outperforms the HDP-HSMM in terms of estimation and consistency. This paper concludes that the rHDP-HSMM is worth applying to datasets where an HDP prior may be generating redundant states. Further inspection as to how to select the threshold may be required, however it is clear that the merging procedure within the model is still able to learn consistent and highly interpretable states for the study of driving maneuvers.



\bibliographystyle{elsarticle-num-names} 
\bibliography{robust_hsmm}       





\end{document}